\documentclass[runningheads]{llncs}
\usepackage{mymacros}
\usepackage[utf8]{inputenc}
\usepackage{booktabs}
\usepackage[english]{babel}
\usepackage{cite}
\usepackage{makeidx}  
\usepackage{amssymb}
\usepackage{amsmath}
\usepackage{mathtools}
\usepackage{nicefrac,xfrac}
\usepackage{graphicx}
\usepackage{multirow}
\usepackage{hyperref}
\usepackage{subcaption}
\usepackage{academicons}
\usepackage{xcolor}
\usepackage{longtable}
\usepackage[flushleft]{threeparttable}
\usepackage{makecell}
\usepackage{tabstackengine}
\TABstackMath
\setstackgap{S}{2pt}

\definecolor{orcidlogocol}{HTML}{A6CE39}

\addto\extrasenglish{%
}

\begin{document}
\title{Detecting Context-Aware Deviations in Process Executions}
%
%
\author{Gyunam Park
\and
Janik-Vasily Benzin
\and
Wil M. P. van der Aalst
}
\authorrunning{Park et al.}
%
\institute{Process and Data Science Group (PADS), RWTH Aachen University, Germany \\
\email{gnpark@pads.rwth-aachen.de, janik.benzin@rwth-aachen.de, wvdaalst@pads.rwth-aachen.de}
}
\maketitle              
\begin{abstract}
A deviation detection aims to detect deviating process instances, e.g., patients in the healthcare process and products in the manufacturing process.
A business process of an organization is executed in various contextual situations, e.g., a COVID-19 pandemic in the case of hospitals and a lack of semiconductor chip shortage in the case of automobile companies.
Thus, \emph{context-aware deviation detection} is essential to provide relevant insights.
However, existing work 1) does not provide a systematic way of incorporating various contexts, 2) is tailored to a specific approach without using an extensive pool of existing deviation detection techniques, and 3) does not distinguish \emph{positive} and \emph{negative} contexts that justify and refute deviation, respectively.
In this work, we provide a framework to bridge the aforementioned gaps.
We have implemented the proposed framework as a web service that can be extended to various contexts and deviation detection methods.
We have evaluated the effectiveness of the proposed framework by conducting experiments using 255 different contextual scenarios.
\end{abstract}
\section{Introduction}
Deviation detection in process executions aims to identify anomalous executions by distinguishing deviating behaviors from normal behaviors.
A range of deviation detection techniques for business processes has been proposed~\cite{Bohmer2017deviationreview}.
The techniques are categorized as supervised and unsupervised ones.
The former defines normal behavior to identify deviations of recorded process executions with respect to the specified normal behavior, whereas the latter identifies deviations without such normal behaviors.
Since many businesses lack the specification of normal behavior, unsupervised deviation detection techniques recently gained more attention~\cite{Bohmer2017deviationreview}.

As a process is executed in a specific \textit{context} (e.g., COVID-19 Pandemic) that affects the behavior of the execution, it is indispensable to consider the context when detecting deviations~\cite{vanderaalst2012context}.
In this regard, \textit{context-aware deviation detection} aims to classify a trace (i.e., a sequence of events by a process instance) to \raisebox{.5pt}{\textcircled{\raisebox{-.9pt} {1}}} \textit{context-insensitive normal} meaning the trace is normal regardless of context, \raisebox{.5pt}{\textcircled{\raisebox{-.9pt} {2}}} \textit{context-insensitive deviating} meaning the trace is deviating regardless of context, \raisebox{.5pt}{\textcircled{\raisebox{-.9pt} {3}}} \textit{context-sensitive normal} meaning the trace is deviating without considering context but normal when considering context, and \raisebox{.5pt}{\textcircled{\raisebox{-.9pt} {4}}} \textit{context-sensitive deviating} meaning the trace is normal without considering context but deviating when considering context.

Few approaches have been developed to (indirectly) solve the context-aware deviation detection problem~\cite{Bohmer2017deviationreview}.
For instance, Pauwels et al.~\cite{Pauwels2019bayesiannetworksdeviaton} extend Bayesian networks to learn conditional probabilities for organizational contexts such as roles of resources.
Warrender et al.~\cite{warrender1999tstide} propose a sliding-window based approach that considers time-related context.
Mannhardt et al.~\cite{Mannhardt2016alignmentsbalanceddeviation} conceptualize context as data attributes of process instances.

However, each approach is tailored to consider limited aspects of contexts, not providing a systematic way to extend the approach to consider various aspects of contexts.
Given a large space of possibly relevant contexts proposed in studies on contexts (cf. \autoref{subsec:related-context}), we need a systematic framework to integrate context to deviation detection.

Moreover, a framework to integrate a large number of existing deviation detection methods with different strengths and weaknesses on varying assumptions is missing.
Instead, the existing work is confined to a single method and inherits the methods’ unique set of properties.

Furthermore, existing techniques do not distinguish \emph{positive} and \emph{negative} contexts.
The former justifies deviations.
For instance, COVID-19 Pandemic in a healthcare process explains the long waiting time for admission, e.g., due to the sudden increase in the number of patients.
The latter refutes non-deviations.
``Crunch time'' in a video game industry denies a normal throughput time of the game development process, e.g., with the compulsory overwork by employees.
Existing work considers only negative contexts when integrating context into deviation detection.

\begin{figure}[ht!]
    \center
    \includegraphics[width=1\linewidth, trim = 0mm 0mm 0mm 0mm, clip]{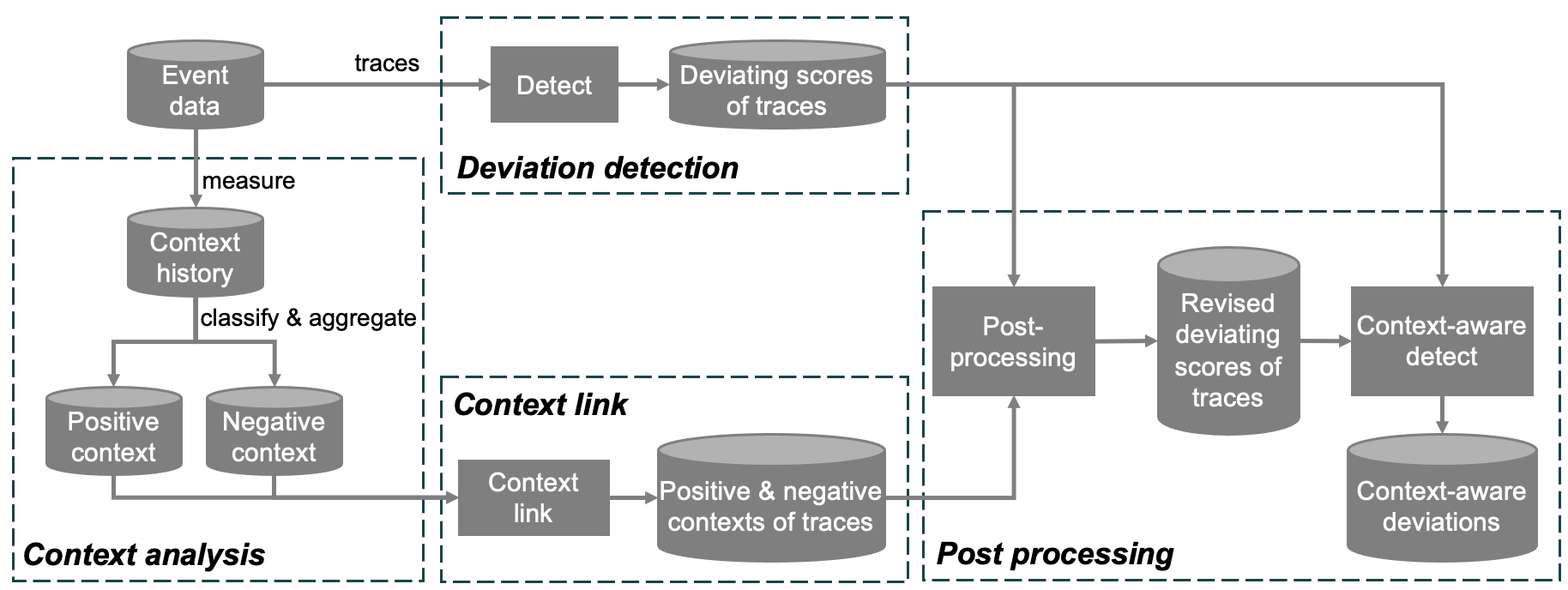}
    \caption{An overview of the framework for context-aware deviation detection}
    \label{fig:framework-overview}
\end{figure}

In this paper, we propose a framework based on \emph{post-processing mechanism} to systematically support the context-aware deviation by integrating the extensive existing deviation detection methods and contexts. 
As shown in \autoref{fig:framework-overview}, the framework consists of four components.
First, \emph{deviation detection} computes deviating scores of traces, with which we can classify \textit{non-context deviating} and \textit{non-context normal} traces.
Next, \emph{context analysis} computes \emph{positive} and \emph{negative} contexts by aggregating \emph{context history}.
Afterwards, \emph{context link} connects the context to traces.
Next, \emph{post-processing} increases the deviation score of a trace with the positive context of the trace and decreases it with the negative context.
Using the revised deviation score, we classify traces as \textit{context-normal} and \textit{context-deviating}.
Finally, we label a trace as one of \raisebox{.5pt}{\textcircled{\raisebox{-.9pt} {1}}}-\raisebox{.5pt}{\textcircled{\raisebox{-.9pt} {4}}} based on the \emph{non-context} and \emph{context} classifications.

To summarize, this paper provides the following contributions:
\begin{itemize}
    \item We propose a framework to solve the context-aware deviation detection problem while integrating the existing deviation detection methods and contexts.
    \item We extend the context conceptualization with \textit{positive} and \textit{negative} contexts that carry dedicated semantics for deviation detection.
    \item We implement a flexible and scalable web service supporting the framework and evaluate the effectiveness of the framework with 225 simulated scenarios.
\end{itemize}

The remainder is organized as follows. We discuss the related work in \autoref{sec:related}. Then, we present the preliminaries in \autoref{sec:preliminaries}. Next, we introduce the context-awareness in \autoref{sec:context} and a framework for integrating contexts and deviation detection in \autoref{sec:framework}. 
Afterward, \autoref{sec:implementation} introduces the implementation of a web application, and \autoref{sec:evaluation} evaluates the effectiveness of the proposed framework.
Finally, \autoref{sec:conclusion} concludes the paper.

\section{Related Work}\label{sec:related}
In this section, we introduce existing literature on unsupervised deviation detection of process executions and the context of business processes.

\subsection{Unsupervised Deviation Detection}\label{subsec:related-deviation}
Unsupervised deviation detection is categorized into 1) process-centric, 2) profile-based, 3) process-agnostic and interpretable, and 4) process-agnostic and non-interpretable methods.

\textbf{Process-centric.}
\cite{Bezerra2013deviationsimplealgorithms} computes the conformance of traces to a process model and classifies non-conforming traces as deviating.
\cite{Bohmer2017deviationwindowbasedsignatureextraction} refines the concept of likelihood graphs by mining small likelihood graph signatures from event data. 
A deviation is determined by comparing the execution likelihood of a trace with respect to a set of mined signatures and a reference likelihood.
\cite{Jalali2010geneticdeviationclassicprocessmodelusinggeneticdiscovery} discovers process models using genetic algorithms and conducts conformance checking using token-based replay to detect deviating traces.

\textbf{Profile-based.}
\cite{Li2017profilesdeviation} iteratively samples more normal sets of traces and \textit{profiles} each trace against the more normal set of traces. 
The result is a sorted list of traces according to their profiles in the last iteration, which is used to partition the event data into a set of normal traces and a set of deviating traces using a deviation threshold. 

\textbf{Process-agnostic and interpretable.}
\cite{Pauwels2019bayesiannetworksdeviaton} extends Bayesian networks and defines a conditional likelihood-based score using the extended bayesian network on traces.
All traces are then sorted according to the score, and the first \emph{k} are returned as deviating traces.
\cite{Bohmer2020deviationadar} uses \emph{association rules}. 
A set of anomaly detection association rules specifying normal behavior is mined from the event data. 
A trace is detected as deviating if its aggregate support is below the aggregate support of its most similar trace in the event data with respect to the set of anomaly detection association rules.
\cite{warrender1999tstide} uses a sliding window-based approach to extract frequency information over those windows. 
If a trace contains infrequent windows, then it is deviating.

\textbf{Process-agnostic and non-interpretable.}
\cite{Nolle2018autoencoderdeviation} encodes traces in event data using one-hot encoding and train \emph{autoencoder neural network} with them.
The deviation of a trace is determined using the error the autoencoder makes in predicting the trace. 
\cite{Nolle2019deviationrnn} further develops the application of neural networks to event data by training a recurrent neural network to predict the next event in integer-encoding based on the current event in a trace. 
The aggregate likelihood of predicting the correct events is used to detect deviations. 

Some of the unsupervised deviation detection methods provide room for handling limited kinds of context but take method-dependent approaches such that neither a general integration nor support for a systematic extension of context is provided.
In this work, we provide a general framework to integrate various unsupervised deviation detection techniques with different strengths, weaknesses, and assumptions to systematically extend them with contexts.

\subsection{Context}\label{subsec:related-context}
In pervasive computing, especially for developing adaptive services,
context is conceptualized as the lower level of the abstraction of raw data~\cite{ye2012situationreview}. Another higher-level abstraction, called situation, is introduced to map one or multiple contexts to semantically richer concepts such as users' behaviors.

In business processes, a context is a multitude of concepts that affect the behavior and performance of the process. 
\cite{vanderaalst2012context} derives four levels of context that should be considered during the analysis of processes to improve the quality of results.
\cite{Song2019contextreview} extends it and provides an ontology of contexts in BPM by conducting an extensive literature review of the context in BPM.

More ontological approaches have been proposed to specify context and situations.
Generally, they categorize contexts into \emph{intrinsic} and \emph{relational}.
\cite{costa2007contextrules} differentiate between \emph{intrinsic} and \emph{relational} context whereby intrinsic context is essential to the nature of the entity and relational context is inherent to the relation of multiple entities.
\cite{Kronsbein2014contextinternal-org-resource-customer-external} develops a two-level framework for structuring context, which is more coarse-grained than the four levels of \cite{vanderaalst2012context}. 

In this work, we merge relevant contexts of the earlier work and their categorizations into an integrative context ontology that is aimed at extracting context from event data.

\section{Preliminaries}\label{sec:preliminaries}
\begin{definition}[Event]
Let $\univ{e}$ be the universe of events,
Let $\univ{att}{=}\{act,case,\allowbreak time,\dots\}$ be the universe of attribute names.
For any $e \in \univ{e}$ and $att \in \univ{att}: \#_{att}(e)$ is the value of attribute $att$ for event $e$, e.g., $\#_{time}(e)$ indicates the timestamp of event $e$.
\end{definition}

\begin{definition}[Trace]
A trace is a finite sequence of events $\sigma \in \univ{e}^*$ such that each event appears only once, i.e., $\forall_{1\le i < j \le |\sigma|} \:\sigma(i)\neq \sigma(j)$.
Given $\sigma \in \univ{e}^*$ and $e \in \univ{e}$, we write $e \in \sigma$ if and only if $\exists_{1 \le i \le |\sigma|}\: \sigma(i){=}e$.
We define $\mi{elem} \in \univ{e}^* \rightarrow \pow(\univ{e})$ with $\mi{elem}(\sigma){=}\{e \in \sigma \}$.
\end{definition}

\begin{definition}[Event Log]
An event log is a set of traces $L \subseteq \univ{e}^*$ such that each event appears at most once in the event log, i.e., for any $\sigma_1,\sigma_2 \in L$ such that $\sigma_1 \neq \sigma_2:elem(\sigma_1) \cap elem(\sigma_2){=}\emptyset$.
Given $L \subseteq \univ{e}^*$, we denote $E(L){=}\bigcup_{\sigma \in L} elem(\sigma)$.
\end{definition}

\begin{definition}[Time Window]
Let $\univ{time}$ be the universe of timestamps.
$\univ{tw}=\{(t_s,t_e \in \univ{time} \times \univ{time} \mid t_s \le t_e \}$ is the set of all possible time windows.
$\mi{duration} \in \univ{tw} \rightarrow \mathbb{R}$ maps a time window to a real valued representation of the difference between the its start and end in the granularity of seconds.
\end{definition}

For $tw {=} (t_{s},t_{e})$, $\pi_{s}(tw) {=} t_{s}$ and $\pi_{e}(tw) {=} t_{e}$.
For instance, $tw_1{=}(\text{\scriptsize 2022-01-01 00:00:00}, \allowbreak \text{\scriptsize 2022-01-08 00:00:00})$ is a time window where $\pi_{s}(tw_1){=} \text{\scriptsize 2022-01-01 00:00:00}$, $\pi_{e}(tw_1){=}\text{\scriptsize 2022-}\\ \text{\scriptsize 01-08 00:00:00}$, and $\mi{duration}(tw_1){=}604800$ (seconds).
Note that, in the remainder, we denote $604800$ as $\mi{week}$.

A time span of an event log with length $l$ is a collection of non-overlapping time windows of the event log that have the equal duration of $l$.

\begin{definition}[Time Span]
Let $l \in \mathbb{R}$ be a time span length.
Let $L \in \univ{e}^*$ be an event log.
$t_{\mi{min}}(L)=min_{e \in E(L)} \#_{time}(e)$, $t_{\mi{max}}(L)=max_{e \in E(L)} \#_{time}(e)$, and $n_{l}(L)=\lceil \nicefrac{(t_\mi{max}(L) - t_{\mi{min}(L)})}{l} \rceil$.
$\mi{span}_{l}(L)=\{(t_{\mi{min}}(L)+(k-1) \cdot l, t_{\mi{min}}(L) + k \cdot l) \mid 1 \le k \le n_{l}(L) \}$.
For any $e \in E(L)$, $tw_{l,L}(e)=tw$ s.t. $\pi_{s}(tw) \le \#_{time}(e) \le \pi_{c}(tw)$.
\end{definition}

Assume that event log $L$ contains traces that consist of events between \text{\scriptsize 2022-01-01 00:00:00} and \text{\scriptsize 2022-01-15 00:00:00}.
$t_{\mi{min}}(L)=\text{\scriptsize 2022-01-01 00:00:00}$, $t_{\mi{max}}(L)=\text{\scriptsize 2022-01-15 00:00:00}$, and $n_{\mi{week}}(L)=2$.
$\mi{span}_{\mi{week}}(L)$ contains two time windows $tw_1 \allowbreak = (\text{\scriptsize 2022-01-01 00:00:00},\text{\scriptsize 2022-01-08 00:00:00})$ and $tw_2{=}(\text{\scriptsize 2022-01-08 00:00:00},\text{\scriptsize 2022-01-15 00:00:00})$.
\section{Context-Aware Deviation Detection}\label{sec:context}
In this section, we introduce a context-aware deviation detection problem and explain an ontology of contexts for context-aware deviation detection.

\subsection{Context-Aware Deviation Detection Problem} \label{ssec:framework-problem}
First, a deviation detection problem is to compute a function that labels traces either with label \textit{deviating} or with label \textit{normal}.
All known deviation detection methods implicitly or explicitly use some form of scoring of traces \emph{score} that is a mapping of traces to some real number (cf. \autoref{subsec:related-deviation}). 
A threshold $\tau$ is used to decide the label.
We conceptualize deviating traces as traces scored above $\tau$.

\begin{definition}[Deviation Detection]\label{def:deviation-detection}
    Let $L$ be an event log.
    Let $\mathbb{S}{=}[0,1]$ be a range of all possible score values and $\tau \in \mathbb{S}$ be a threshold value.
    A score function $\mathit{score} \in L \rightarrow \mathbb{S}$ maps traces to score values.
    $\mi{detect}_{\mi{score}} \in L \rightarrow \{ d, n \}$ is a deviation detection using $score$ such that, for any $\sigma \in L$, $\mi{detect}_{\mi{score}}(\sigma)=\mi{d}$ if $\mi{score}(\sigma)>\tau$. $\mi{detect}_{\mi{score}}(\sigma)=\mi{n}$ otherwise.
\end{definition}

Instead of the two-class labeling problem, a context-aware deviation detection problem is a four-class labeling problem.
\autoref{tab:context-aware-problem} describes the four classes with two dimensions: \textit{non-context} and \textit{context}.
The non-context deviating ($d$) and normal ($n$) correspond to the two classes of the deviation detection problem, whereas context-deviating ($d_c$) and context-normal ($n_c$) indicate that a trace is deviating and normal, respectively, when considering context.
First, \emph{context-insensitive deviating} (i.e., $d {\Rightarrow} d_c$) indicates that a trace is both non-context deviating and context-deviating.
Second, \emph{context-sensitive deviating} (i.e., $n {\Rightarrow} d_c$) denotes that a trace is non-context normal, but context-deviating.
Third, \emph{context-sensitive normal} (i.e., $d {\Rightarrow} n_c$) indicates that a trace is non-context deviating, but context-normal.
Finally, \emph{context-insensitive normal} (i.e., $n {\Rightarrow} n_c$) denotes that a trace is both non-context normal and context-normal.

\begin{table}[]
\centering
\caption{Four classes in a context-aware deviation detection problem}\label{tab:context-aware-problem}
\begin{tabular}{|cc|cc|}
\hline
\multicolumn{2}{|c|}{\multirow{2}{*}{$\sigma \in \univ{e}^*$}} & \multicolumn{2}{c|}{\textbf{Context}}
\\
\cline{3-4} 
\multicolumn{2}{|c|}{} & \multicolumn{1}{c|}{Deviating ($d_c$)} & Normal ($n_c$)
\\ 
\hline
\multicolumn{1}{|c|}{\multirow{2}{*}{\textbf{Non-context}}} & \begin{tabular}[c]{@{}c@{}}Deviating
\\
($d$)\end{tabular} & \multicolumn{1}{c|}{\begin{tabular}[c]{@{}c@{}}Context-insensitive deviating\\($d {\Rightarrow} d_c$)\end{tabular}} & \begin{tabular}[c]{@{}c@{}}Context-sensitive normal\\($d {\Rightarrow} n_c$)\end{tabular}
\\
\cline{2-4} 
\multicolumn{1}{|c|}{} & \begin{tabular}[c]{@{}c@{}}Normal
\\
($n$)\end{tabular} & \multicolumn{1}{c|}{\begin{tabular}[c]{@{}c@{}}Context-sensitive deviating
\\($n {\Rightarrow} d_c$)\end{tabular}} & \begin{tabular}[c]{@{}c@{}}Context-insensitive normal\\ ($n {\Rightarrow} n_c$)\end{tabular}
\\
\hline
\end{tabular}
\end{table}

\begin{definition}[Context-Aware Deviation Detection Problem] \label{def:deviation-problem-context}
    Given $L \subseteq \univ{e}^*$, compute a function that labels traces with context-insensitive deviating, context-sensitive deviating, context-sensitive normal, or context-insensitive normal, i.e., $\mi{c\mhyphen detect} \in L \rightarrow \{ d {\Rightarrow} d_c, n {\Rightarrow} d_c, d {\Rightarrow} n_c, n {\Rightarrow} n_c \}$.
\end{definition}

\subsection{Context-Awareness}
\begin{figure}[ht!]
    \center
    \includegraphics[width=0.6\linewidth, trim = 0mm 0mm 0mm 0mm, clip]{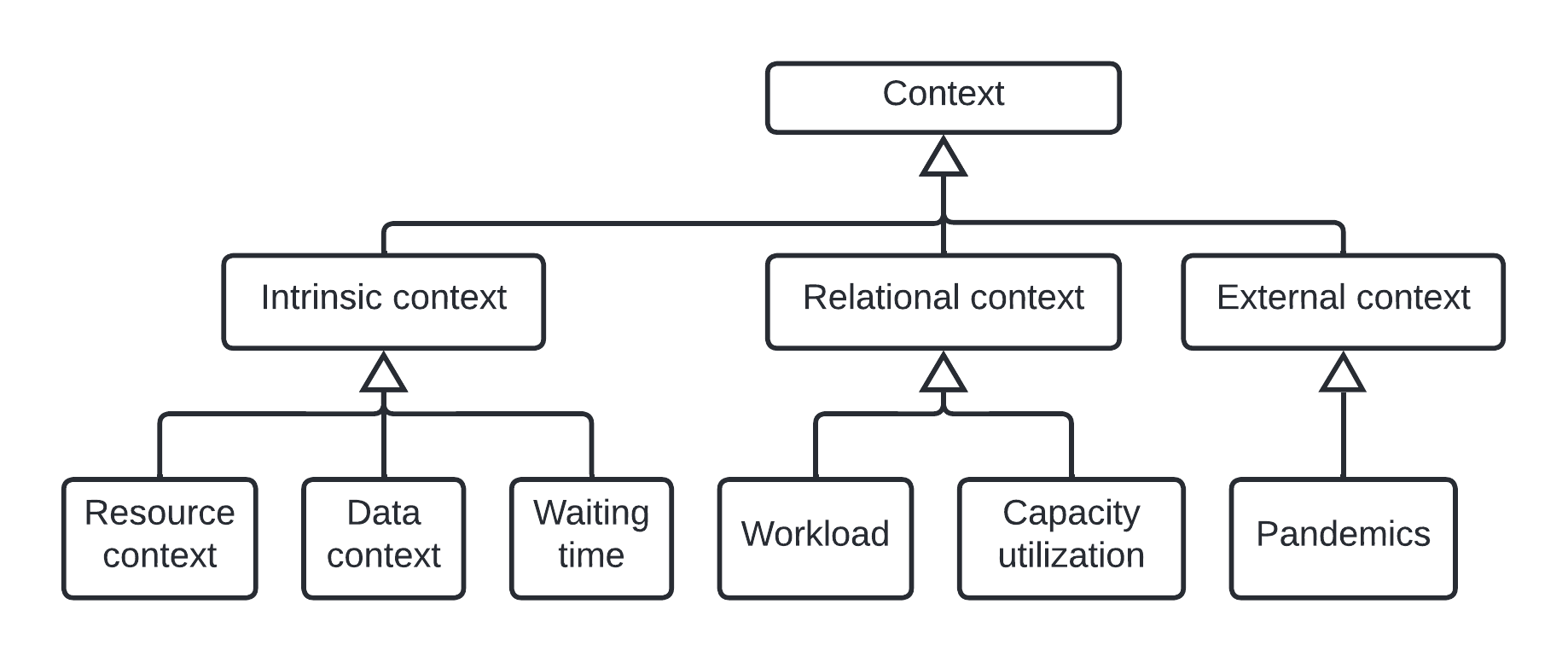}
    \caption{An ontology of business process context for deviation detection \cite{vanderaalst2012context}  \cite{Song2019contextreview} \cite{costa2007contextrules} \cite{Kronsbein2014contextinternal-org-resource-customer-external}.}
    \label{fig:context-ontology}
\end{figure}

Based on existing work on contexts of business processes introduced in \autoref{subsec:related-context}, we provide context ontology for context-aware deviation detection in \autoref{fig:context-ontology}.
First, \emph{intrinsic} context is inherent to an event.
The intrinsic contexts \emph{resource} and \emph{data} correspond to the organizational and data perspectives for a single event.
The \emph{waiting time} context represents the average waiting time of an event. 
Thus, the information of \emph{waiting time} contexts can be used to capture unusually long delays for events.

Next, \emph{relational} context is inherent to the relation of multiple events.
The relational contexts \emph{workload}, \emph{waiting time} and \emph{capacity utilization} represent context information that is measured (extracted) by relating multiple events of the data. 
The \emph{workload} context represents event counts of various selections for a given time window.
The \emph{capacity utilization} context represents workloads of resources or locations of events by counting the respective events that were recorded during the time window of the context, e.g., the capacity utilization of a finance department. 
Therefore, the information of \emph{capacity utilization} contexts can be used to capture unusually high workloads of resources.

Finally, \emph{external} context is not directly attributable to events, but still affects them. 
The external context \emph{pandemics} represents the outbreak of infectious disease, e.g., COVID-19 pandemic.
As an external context is not directly measurable on event data, either additional data has to be used, or it has to be represented by another measurable relational context caused by the external context, e.g., a hygienic products shop experiences exceptionally large demand during the first worldwide outbreak of Corona pandemic such that the \emph{workload} context captures the unusual demand increase and, thus, the external context \emph{pandemic}.

\section{Framework for Context-Aware Deviation Detection}\label{sec:framework}
This section introduces a framework based on \emph{post-processing mechanism}. 
We explain each of the four components described in \autoref{fig:framework-overview} with a running example: 1) deviation detection, 2) context analysis, 3) context link, and 4) post processing.

\subsection{Running Example}
\begin{figure}[ht!]
    \center
    \includegraphics[width=0.9\linewidth]{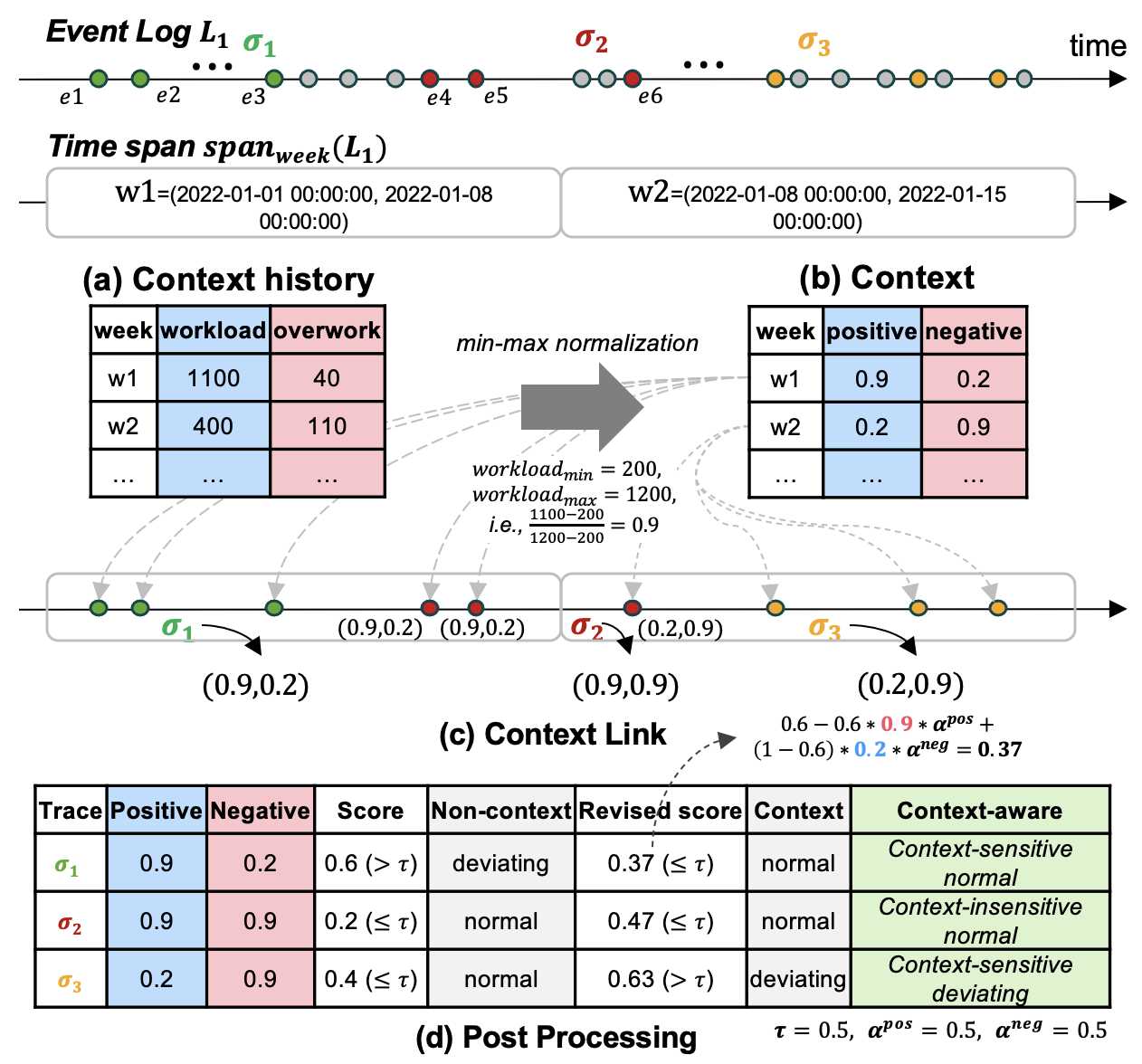}
    \caption{
    A running example of context-aware deviation detection for the time window \textit{week 1 (w1)} and \textit{week 2 (w2)}. 
    \textbf{(a)} The context history of $L_1$ in \textit{w1} shows \textit{workload} of $1100$ (total number of events in \textit{w1}) and \textit{overwork} of $40$ (total number of events during weekend in \textit{w1}), respectively. 
    \textbf{(b)} Assume \textit{workload} is a positive measure, $\mi{workload}_{\mi{max}}{=}1200$, and $\mi{workload}_{\mi{min}}{=}200$.
    By aggregating positive (blue) and negative (red) measures in \textit{w1} with \textit{min-max normalization}, we compute the context in \textit{w1}, i.e., positive context of $0.9$ and negative context of $0.2$.
    \textbf{(c)} We first connect the context to events (as denoted by gray dotted lines) and then connect the context to a trace by computing the maximum positive and negative contexts of its events.
    $\sigma_2$ has the positive context of $0.9$ (i.e., the maximum positive context of its events) and the negative context of $0.9$ (i.e., the maximum negative context).
    \textbf{(d)} The non-context deviating score of $\sigma_1$ is $0.6$ ($> \tau$, i.e., non-context deviating), but its revised deviation score is $0.37$ ($\le \tau$, i.e., context-normal).
    Thus, $\sigma_1$ is context-sensitive normal.
    }
    \label{fig:running-example}
\end{figure}

\autoref{fig:running-example} shows a running example of an order management process.
It describes events of the process for two weeks under 1) the context of high workload (i.e., many events during the week) in \textit{week 1} and 2) the context of overwork (i.e., many events during the weekend) in \textit{week 2}.
The context of high workload is considered as a positive context, i.e., the context justifies deviating traces in \textit{week 1}, producing more context-normal traces.
In contrast, we consider the context of overwork as a negative context, i.e., the context refutes normal traces in \textit{week 2}, producing more context-deviating traces in \textit{week 2}.

\subsection{Context Analysis}
We analyze context in two steps.
First, we compute \textit{context history} based on event logs.
A \emph{context history} describes the value of different measures (e.g., \textit{workload} and \textit{overwork}) in different time windows.

\begin{definition}[Context History] \label{def:measure}
Let $\univ{\mi{measure}}{=}\{\mi{workload}, \mi{overwork}, \dots\}$ be the universe of measure names.
$\univ{\mi{ch}}=\univ{tw} \nrightarrow (\univ{\mi{measure}} \nrightarrow \mathbb{R})$ is the universe of context history.
Let $L$ be an event log and $l \in \mathbb{R}$ a time span length.
$\mi{ch}_{l}(L) \in \univ{\mi{ch}}$ is the context history in $L$ with time span of $l$.
\end{definition}

\autoref{fig:running-example}(a) shows the context history of $L_1$ with time span length $\mi{week}$, i.e., $\mi{ch}_{\mi{week}}(L_1)$.
It contains the measures of \textit{workload} and \textit{overwork}.
For instance, $\mi{ch}_{\mi{week}}(L_1)(\mi{w1})(\mi{workload})=1100$ and $\mi{ch}_{\mi{week}}(L_1)(\mi{w1})\allowbreak(\mi{overwork})=40$.

A context consists of \emph{positive} and \emph{negative} context scores.
They describe the overall positive/negative contexts in a time window with a value ranging from $0$ to $1$, respectively.
The closer the value is to 1, the stronger the respective context is.
We compute the context in a time window using context measures in the context history of the time window.
To this end, we 1) normalize context measures in the time window, 2) distinguish positive and negative context measures, 3) aggregate positive and negative context measures with different weights (i.e., the importance of measures).

\begin{definition}[Context] \label{def:measure}
    Let $L$ be an event log and $l \in \mathbb{R}$ a time span length.
    $\mi{type} \in \univ{\mi{measure}} \rightarrow \{\mi{pos},\mi{neg}\}$ maps measures to $\mi{pos}$ and $\mi{neg}$,
    $\mi{w} \in \univ{\mi{measure}} \rightarrow \mathbb{R}$ maps measures to weights, and $\mi{norm} \in \univ{\mi{measure}} \rightarrow (\mathbb{R} \rightarrow [0,1])$ maps measures to normalization functions that assign values ranging from 0 to 1 to measure values.
    $ctx_{l,L} \in span_{l}(L) \nrightarrow [0,1]^{2}$ is a context such that, for any $tw \in dom(ctx_{l,L})$, $ctx_{l,L}(tw)=(pc,nc)$ with
    \begin{itemize}
        \item $pc = \sum\nolimits_{m \in dom(\mi{ch}_{l,L}^{tw}) \wedge \mi{type}(m) {=} \mi{pos}} \nicefrac{\mi{w}(m) \cdot \mi{norm}(m)(\mi{ch}_{l,L}^{tw}(m))}{\mi{w}(m)}$ and
        \item $nc = \sum\nolimits_{m \in dom(\mi{ch}_{l,L}^{tw}) \wedge \mi{type}(m) {=} \mi{neg}} \nicefrac{\mi{w}(m) \cdot \mi{norm}(m)(\mi{ch}_{l,L}^{tw}(m))}{\mi{w}(m)}$
    \end{itemize}
    , where $\mi{ch}_{l,L}^{tw}=\mi{ch}_{l}(L)(tw)$.
\end{definition}

The example in \autoref{fig:running-example} assumes $\mi{norm}_1$, $\mi{type}_1$, and $\mi{w}_1$.
First, $\mi{norm}_1$ uses \textit{min-max normalization} for each measure, e.g., with the maximum \textit{workload} of $1200$, the minimum \textit{workload} of $200$, the maximum \textit{overwork} of $120$, and the minimum \textit{overwork} of $20$.
Moreover, $\mi{type}_1$ classifies \textit{workload} as a positive context measure and \textit{overwork} as a negative context measure, i.e., $\mi{type}_1(\mi{workload}){=}\mi{pos}$ and $\mi{type}_1(\mi{overwork}){=}\mi{neg}$.
Finally, $\mi{w}_1$ assigns the weights of $10$ and $5$ to \textit{workload} and \textit{overwork}, respectively, i.e., $\mi{w}_1(\mi{workload}){=}10$ and $\mi{w}_1(\mi{overwork}){=}5$.

\autoref{fig:running-example}(b) shows context $ctx_{\mi{week},L_1}$.
The positive context in time window $\mi{w1}$ is 
$\nicefrac{\mi{w}_1(\mi{workload}) \cdot \mi{norm}_1(\mi{workload})(1200)}{\mi{w}_1(\mi{workload})}{=}\nicefrac{10 \cdot 0.9}{10}{=}0.9$.
The negative context in $\mi{w1}$ is
$\nicefrac{\mi{w}_1(\mi{overwork}) \cdot \mi{norm}_1(\mi{overwork})(200)}{\mi{w}_1(\mi{overwork})}{=}\nicefrac{5 \cdot 0.2}{5}{=}0.2$.
Note that, in the example, the weight does not play its role since we only use one positive and one negative context measure.

\subsection{Linking Context to Traces}
To connect context to traces, we first link context to events. 
An event is connected to the context of the time window that the event belongs to.

\begin{definition}[Context-Event Link] \label{def:link}
    Let $L$ be an event log and $l \in \mathbb{R}$ a time span length.
    A context-event link, $\mi{elink}_{l,L} \in E(L) \to [0,1]^2$, maps events to positive and negative contexts such that, for any $e \in E(L)$, $\mi{elink}_{l,L}(e){=}\mi{ctx}_{l,L}(tw_{l,L}(e))$.
\end{definition}

As depicted in \autoref{fig:running-example}(c) by gray dotted lines, $e1$, $e2$, and $e3$ by $\sigma_1$ and $e4$ and $e5$ by $\sigma_2$ are connected to $ctx_{\mi{week}, L_1}(\mi{w1})$, i.e., $\mi{elink}_{\mi{week}, L_1}(e1){=}ctx_{\mi{week}, L_1}(\mi{w1}){=}(0.9,\allowbreak 0.2)$, etc.

The context of a trace is determined by the context of its events.
In this work, we define the maximum positive and negative context of the events of a trace as the context of the trace.

\begin{definition}[Context-Trace Link] \label{def:sit-lift}
    Let $L$ be an event log and $l \in \mathbb{R}$ a time span length.
    $\mi{tlink}_{l,L} \in L \rightarrow [0,1]^2$ maps traces to positive and negative contexts s.t., for any $\sigma \in L$, $\mi{tlink_{l,L}}(\sigma) {=} (\max(\{ pc \in [0,1] \mid \exists_{e \in elem(\sigma)}\; (pc,nc){=}\mi{elink_{l,L}}(e)\})\allowbreak, \max(\{nc \in [0,1] \mid \exists_{e \in elem(\sigma)}\; (pc,nc){=}\mi{elink_{l,L}}(e)\}))$.
\end{definition}

As shown in \autoref{fig:running-example}(c), 
$\sigma_1$ has the positive context of $0.9$ and negative context of $0.2$, i.e., 
$\mi{tlink}_{\mi{week}, L_1}(\sigma_1){=} (0.9, 0.2)$, since the maximum positive context of its events, i.e., $e_1$, $e_2$, and $e_3$, is $0.9$ and the maximum negative context is $0.2$.
$\mi{tlink}_{\mi{week}, L_1}(\sigma_2){=} (0.9, 0.9)$, since the maximum positive context of its events, i.e., $e_4$, $e_5$, and $e_6$, is $0.9$ and the maximum negative context is $0.9$.

\subsection{Post Processing}

Post-processing function revises the non-context deviating score of a trace using the positive and negative context of the trace.
The positive context decreases the deviating score, whereas the negative context increases it.

\begin{definition}[Post Processing] \label{def:post-candidate}
    Let $L$ be an event log, $l \in \mathbb{R}$ a time span length, and $\mi{score}$ a score function.
    $\mi{post_{l,L,\mi{score}}} \in L \times [0,1]^2 \rightarrow [0,1]$ maps a trace, a positive degree, and a negative degree to revised score such that, for any $\sigma \in L$, $\alpha^{\mi{pos}} \in [0,1]$, and $\alpha^{\mi{neg}} \in [0,1]$, $\mi{post_{l,L,\mi{score}}}(\sigma, \alpha^{\mi{pos}}, \alpha^{\mi{neg}}) {=} \mi{score}(\sigma) - \mi{score}(\sigma) \cdot \alpha^{\mi{pos}} \cdot \mi{pc} + (1 - \mi{score}(\sigma)) \cdot  \alpha^{\mi{neg}} \cdot \mi{nc}$ where $(\mi{pc}, \mi{nc}){=}\mi{tlink}_{l,L}(\sigma)$.
\end{definition}

In \autoref{fig:running-example}(d), $\sigma_1$ has the deviation score of $0.6$, i.e., $\mi{score_1}(\sigma_1) {=} 0.6$.
Given $\sigma_1$, $\alpha^{\mi{pos}}{=}0.5$ and $\alpha^{\mi{neg}}{=}0.5$, $\mi{post_{\mi{week}, L_1,\mi{score_1}}}$ revises the deviating score to a new score of $0.37$, i.e., $0.6-0.6\cdot0.5\cdot0.9+(1-0.6)\cdot0.5\cdot0.2{=}0.37$.

Finally, a context-aware detection function labels traces with the four context-aware classes described in \autoref{tab:context-aware-problem}, based on the non-context deviating score and revised deviating score.

\begin{definition}[Context-Aware Detection] \label{def:aware}
    Let $L$ be an event log and $l \in \mathbb{R}$ a time span length.
    Let $\mi{score}$ be a score function.
    Let $\alpha^{\mi{pos}},\alpha^{\mi{neg}} \in [0,1]$ be positive and negative degrees and $\tau \in \mathbb{S}$ be a threshold.
    $\mi{c\mhyphen detect}  \in L \rightarrow \{ d {\Rightarrow} d_c, n {\Rightarrow} d_c, d {\Rightarrow} n_c, n {\Rightarrow} n_c \}$ maps traces to context-aware labels such that for any $\sigma \in L$:
            \[   
                \mi{c\mhyphen detect} (\sigma) {=} 
                     \begin{cases}
                     d {\Rightarrow} d_c &\text{if $\mi{detect}_{\mi{score}}(\sigma) {=} d $ and $\mi{post}_{l,L,\mi{score}}(\sigma,\alpha^{\mi{pos}}, \alpha^{\mi{neg}})$}>\text{$\tau$}\\
                     n {\Rightarrow} d_c &\text{if $\mi{detect}_{\mi{score}}(\sigma) {=} n $ and $\mi{post}_{l,L,\mi{score}}(\sigma,\alpha^{\mi{pos}}, \alpha^{\mi{neg}})$}>\text{$\tau$}\\
                     d {\Rightarrow} n_c &\text{if $\mi{detect}_{\mi{score}}(\sigma) {=} d $ and $\mi{post}_{l,L,\mi{score}}(\sigma,\alpha^{\mi{pos}}, \alpha^{\mi{neg}})$}\le\text{$\tau$}\\
                     n {\Rightarrow} n_c &\text{if $\mi{detect}_{\mi{score}}(\sigma) {=} n $ and $\mi{post}_{l,L,\mi{score}}(\sigma,\alpha^{\mi{pos}}, \alpha^{\mi{neg}})$}\le\text{$\tau$}\\
                     \end{cases}
                \]
\end{definition}

As shown in \autoref{fig:running-example}(d), given $\tau {=} 0.5$, $\alpha^{\mi{pos}} {=} 0.5$, and $\alpha^{\mi{neg}} {=} 0.5$,
$\mi{c\mhyphen detect} (\sigma_1) {=} \allowbreak d {\Rightarrow} n_c$ since $\mi{detect}_{\mi{score_1}}(\sigma_1){=}d$ and $\mi{post}_{\mi{week},L_1,\mi{score_1}}(\sigma_1,\alpha^{\mi{pos}}, \alpha^{\mi{neg}}){=}0.37 \le \tau$.
Furthermore, $\mi{c\mhyphen detect} (\sigma_3) {=} n {\Rightarrow} d_c$ since $\mi{detect}_{\mi{score_1}}(\sigma_2) {=} n$ and $\mi{post}_{\mi{week},L_1,\mi{score_1}}(\allowbreak \sigma_2, \alpha^{\mi{pos}}, \alpha^{\mi{neg}}){=}0.63 > \tau$.
\section{Implementation}\label{sec:implementation}

The framework for context-aware deviation detection is implemented as a cloud-based web service with a dedicated user interface. 
The implementation is available at \url{https://github.com/janikbenzin/contect} along with the source code, a user manual, and a demo video.
It consists of four functional components: (1) context analysis, (2) deviation detection, (3) context-aware deviation detection, and (4) visualization. 

\begin{figure}[ht!]
    \center
    \includegraphics[width=0.8\linewidth, trim = 0mm 0mm 0mm 0mm, clip]{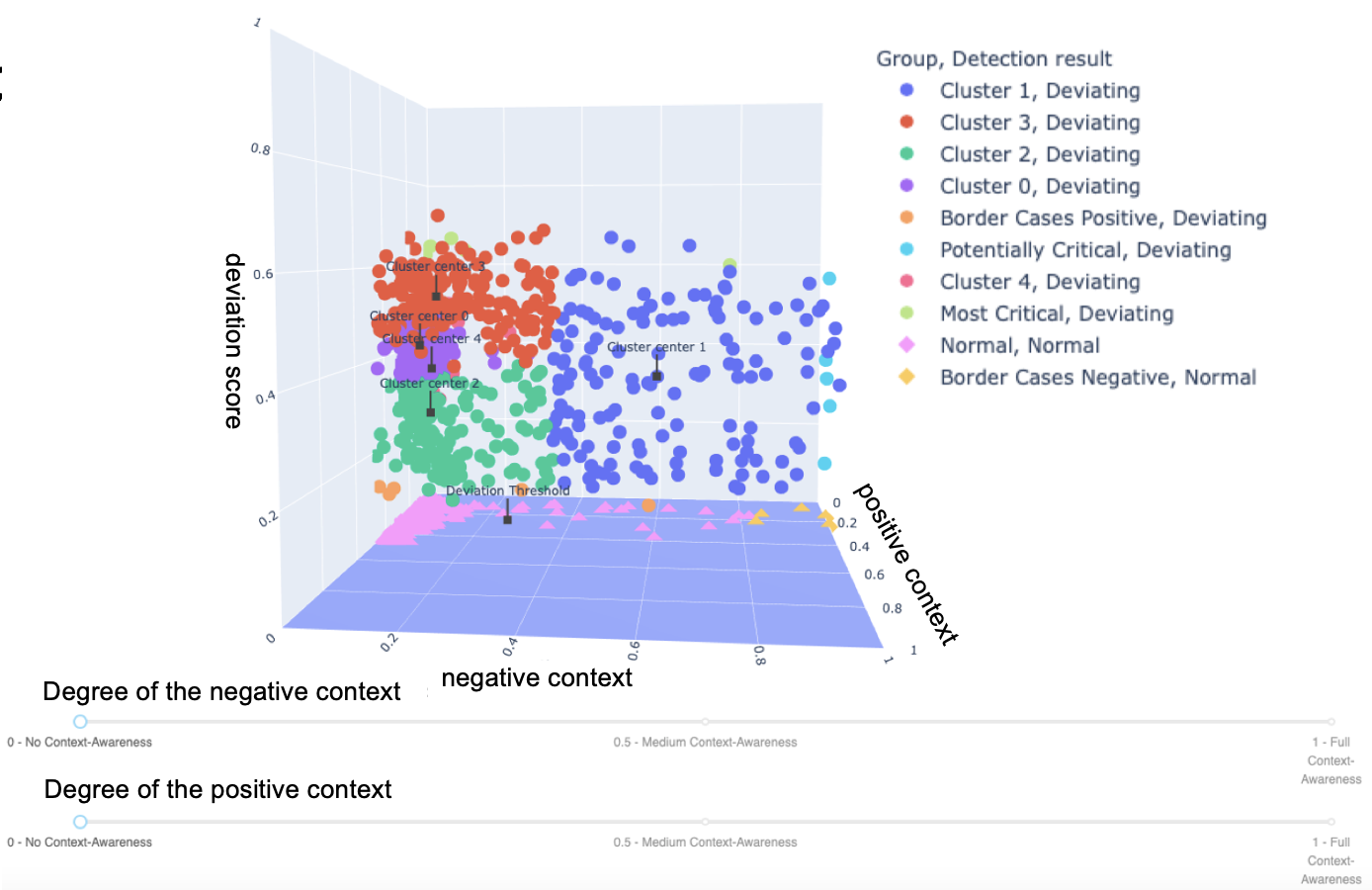}
    \caption{A screenshot of \textit{Scatter} visualization. 
    By varying the degree of positive and negative context, we can deduce the adequate degree of positive and negative context to be used for the context-aware deviation detection.}
    \label{fig:clusters}
\end{figure}

First, the context analysis component supports the computation of the context history and context.
The context introduced in \autoref{fig:context-ontology} have been implemented including \textit{workload}, \textit{weekend}, \textit{waiting time}, and \textit{capacity utilization}.

Second, the deviation detection component implements four deviation detection methods that correspond to representatives of four respective categories introduced in \autoref{subsec:related-deviation}. 
For process-centric methods, we adapt the two-step approach in~\cite{Jalali2010geneticdeviationclassicprocessmodelusinggeneticdiscovery} by using \emph{Inductive} miner~\cite{van2013imf} for process discovery and \emph{alignment}~\cite{vanderaalst2012alignments} for conformance checking.
For profile-based approaches, \emph{Profiles}~\cite{Li2017profilesdeviation} has been implemented, while \emph{ADAR}~\cite{Bohmer2020deviationadar} and \emph{Autoencoder}~\cite{Nolle2018autoencoderdeviation} have been implemented as process-agnostic \& interpretable/non-interpretable approaches, respectively.
Next, the context-aware deviation detection component implements the post processing and the context-aware deviation detection function.

Finally, the visualization component supports an analysis view for each deviation detection method.
Each analysis view consists of three visualizations: \textit{tabular}, \textit{scatter}, and \textit{calendar}.
\textit{Tabular} visualizes the most deviating traces by sorting them based on the deviation score, the proximity to being relabelled as context-normal, etc.
\textit{Scatter} shows a 3D-scatter plot of the deviation score, positive context, and negative context, as shown in \autoref{fig:clusters}.
As the number of deviating traces can be large, the k-Mediods clustering algorithm is applied to all deviating traces such that the user can analyze the medoids to understand the whole space of deviating traces more efficiently (depicted as first to fourth and seventh legend entry in \autoref{fig:clusters}). 
Moreover, by varying the positive and negative degrees, we can analyze the effect of the context on the deviation detection.
\textit{Calendar} visualizes the context over time by aggregating contexts by time and plotting them over the time span.

\section{Evaluation}\label{sec:evaluation}
This section evaluates the proposed framework using the implementation in \autoref{sec:implementation}.
To this end, we conduct four case studies using deviation detection methods: \emph{Inductive}, \emph{Profiles}, \emph{ADAR}, and \emph{Autoencoder}.
In each case study, we compare the performance of context-aware deviation detection and context non-aware deviation detection in 225 different simulated scenarios.
In the rest of this section, we first introduce a detailed experimental design and then report the results.

\subsection{Experimental Design} \label{ssec:evaluation-design}
\begin{figure}[ht!]
    \center
    \includegraphics[width=\linewidth, trim = 4mm 0mm 4mm 0mm, clip]{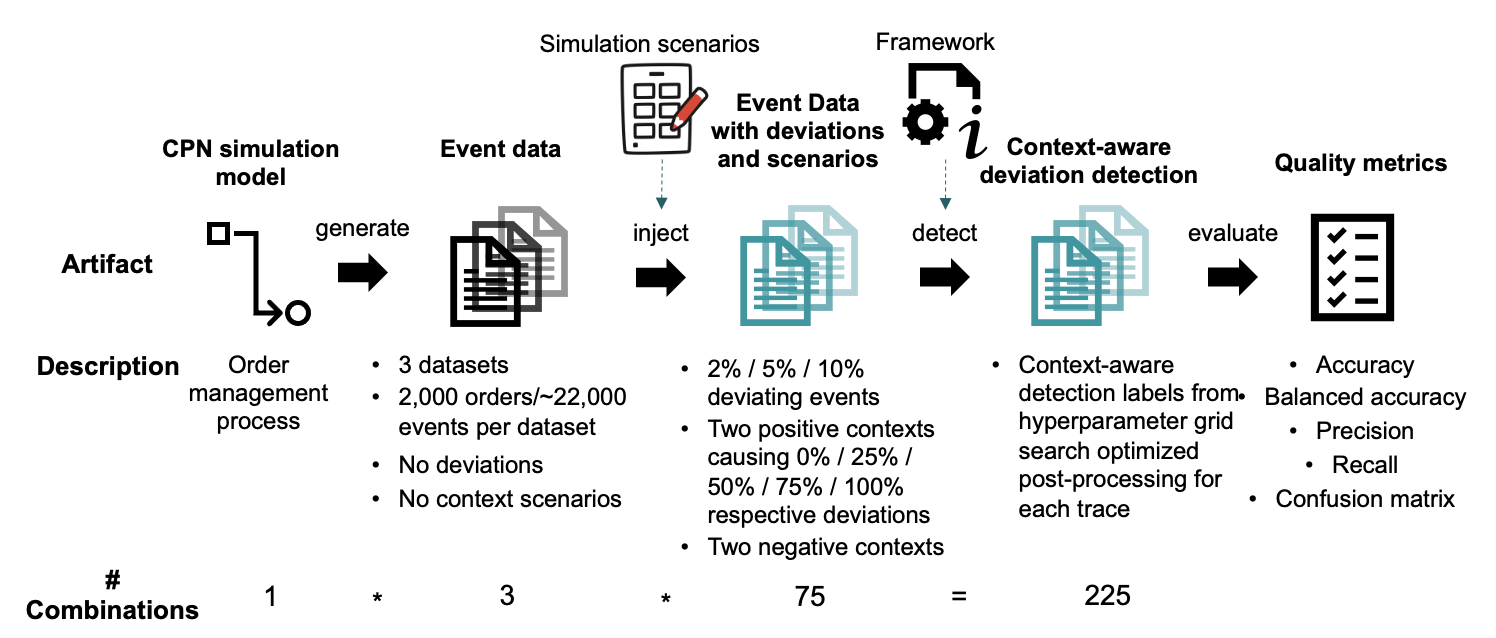}
    \caption{An overview of the experimental design}
    \label{fig:evaluation-design}
\end{figure}

As depicted in \autoref{fig:evaluation-design}, the evaluation follows a four step pipeline: \emph{data generation}, \emph{simulation scenario injection}, \emph{framework application}, and \emph{evaluation of results}.

First, the data generation uses CPN Tools\footnote{\url{www.cpntools.org}} to simulate an order management process.
Next, we inject four different types of deviating events into the generated event data and label them as non-context deviating: 
1) \emph{Rework} randomly adds an event to a trace with the activity that has already occurred, 
2) \emph{Swap} randomly swaps the timestamp of two existing events,
3) \emph{Replace resource} randomly replaces the resource of an event with a different resource, and
4) \emph{Remove} randomly removes an existing event from the data.
To understand the effect of the amount of deviations on the classification result, the evaluation injected 2\%, 5\%, or 10\% deviations equally distributed among the four types. 

Afterward, we inject four contextual scenarios as follows.
\begin{enumerate}
    \item For \emph{workload} scenario, we randomly select a week and add additional orders in the week. 
    We consider it as a positive context and, thus, the non-context deviating events of the selected week are relabelled to context-normal.
    \item For \emph{capacity utilization performance} scenario, we randomly assign vacations and sick leaves to resources, lowering the capacity of the process.
    It is considered as a positive context, and non-context deviating events associated with the reduced capacity resource are relabelled to context-normal.
    \item For \emph{waiting time}, all events of randomly chosen days are randomly delayed. It is considered a negative context, and all of the delayed events that are non-context normal or context-normal are labeled as context-deviating.
    \item For \emph{overwork} scenario, we shift the random percentage of events during weekdays to Saturday and Sunday. It is regarded as a negative context, and all shifted events that are non-context normal or context-normal are relabelled to context-deviating.
\end{enumerate}

To determine the strength of the relationship between positive contexts and deviations, we use \emph{\% context attributable} parameter that determines how many traces are affected by positive contextual scenarios, i.e., non-context deviating events are relabelled to context-normal. 
We include it as the second parameter for experiments with values ranging from 0\% to 100\% as depicted in \autoref{fig:evaluation-design}. 

225 experiments per case study ($3 * 3 * 5 * 5$) result from the parameters as shown in \autoref{fig:evaluation-design}, i.e., three event datasets, three \% events deviating parameters and the five \% context attributable parameters per positive contextual scenario.

Next, we apply the proposed framework and compute context-aware detection results.
Hyperparameter grid search is applied to find the best combination of positive and negative degrees for the post function.

\subsection{Experimental Results} \label{ssec:evaluation-results}
First, we report average results for each case study in \autoref{tab:aggregate-results}, showing that the consideration of positive/negative context is effective in the context-aware deviation detection. 
The first column in \autoref{tab:aggregate-results} shows the performance of context-non-aware deviation detection with $\alpha^{\mi{pos}}$ and $\alpha^{\mi{neg}}$ both set to 0.
The second column in \autoref{tab:aggregate-results} shows the performance of context-aware deviation detection with positive $\alpha^{\mi{pos}}$ and negative degree $\alpha^{\mi{neg}}$ both optimized through the hyperparameter grid search.
The third column shows the performance difference of the proposed approach with respect to the baseline.

In the case study using \emph{Inductive}, the accuracy of \emph{0.389118} is improved by \emph{0.037728} to \emph{0.426846}, the average class accuracy of \emph{0.326856} is slightly reduced by \emph{0.015024} to \emph{0.311832}, the precision of \emph{0.248691} is boosted by \emph{0.044805} to \emph{0.293496} and the recall of \emph{0.389118} is upgraded by \emph{0.037728} to \emph{0.425686}.
The other three case studies also show performance improvements in terms of accuracy, precision, and recall similar to \emph{Inductive} and a decrease in average class accuracy.
In particular, the results are significantly more precise with the framework's context-aware deviation detection than for deviation detection.

\begin{table}[t!]
    \centering
    \caption{Evaluation results from four case studies}
    \label{tab:aggregate-results}
    \begin{threeparttable}
\resizebox{0.8\linewidth}{!}{
\begin{tabular*}{\linewidth}{l l c c c}
    \toprule
    \makecell{} & \makecell{} &  \makecell{Context-non-aware\\deviation detection\\$\alpha^{\mi{pos}} = \alpha^{\mi{neg}} = 0$} &  \makecell{Context-aware\\deviation detection\\$\alpha^{\mi{pos}}, \alpha^{\mi{neg}}$ optimized} & \makecell{Difference}\\
    \midrule
        \multirow{4}{*}{Inductive} & Accuracy &                                                     0.389118 &                     0.426846 & \textcolor{white}{-}0.037728\\
                                    & Avg. class accuracy &                                             0.326856 &                      0.311832 & -0.015024\\
                                    & Precision &                                                       0.248691 &                     0.293496 & \textcolor{white}{-}0.044805\\
                                    & Recall &                                                       0.389118 &                     0.426846 & \textcolor{white}{-}0.037728\\
                                    \multirow{4}{*}{Autoencoder}& Accuracy &                                                         0.385035 &                     0.425686 & \textcolor{white}{-}0.040651 \\
                                    & Avg. class accuracy &                                              0.311249 &                     0.312451 & \textcolor{white}{-}0.001202\\
                                    & Precision &                                                   0.235668 &                     0.369101 & \textcolor{white}{-}0.133433\\
                                    & Recall &                                                       0.385035 &                     0.424996 & \textcolor{white}{-}0.039961\\
            \multirow{4}{*}{\emph{Profiles}}       &   Accuracy &                                                     0.363995 &                     0.406368 & \textcolor{white}{-}0.042373\\
                                    &   Avg. class accuracy &                                            0.293880 &                     0.292083 & -0.001797\\
                                    & Precision &                                                       0.220972 &                     0.332658 & \textcolor{white}{-}0.111686\\
                                    & Recall &                                                      0.363995 &                     0.404011 & \textcolor{white}{-}0.034061\\
                                   
        \multirow{4}{*}{\emph{ADAR}}       & Accuracy &                                                      0.351544 &                     0.395066 & \textcolor{white}{-}0.043522 \\
                                    & Avg. class accuracy       &                                         0.291969 &                     0.289284 & -0.002685\\
                                    & Precision &                                                         0.229760 &                     0.334021 & \textcolor{white}{-}0.104261\\
                                    & Recall &                                                            0.351544 &                     0.385152 & \textcolor{white}{-}0.033608\\
        
    \bottomrule
    \end{tabular*}
}
\end{threeparttable}
\end{table}

\begin{figure}[ht!]
    \begin{subfigure}{.5\textwidth}
        \center
        \includegraphics[width=0.9\linewidth]{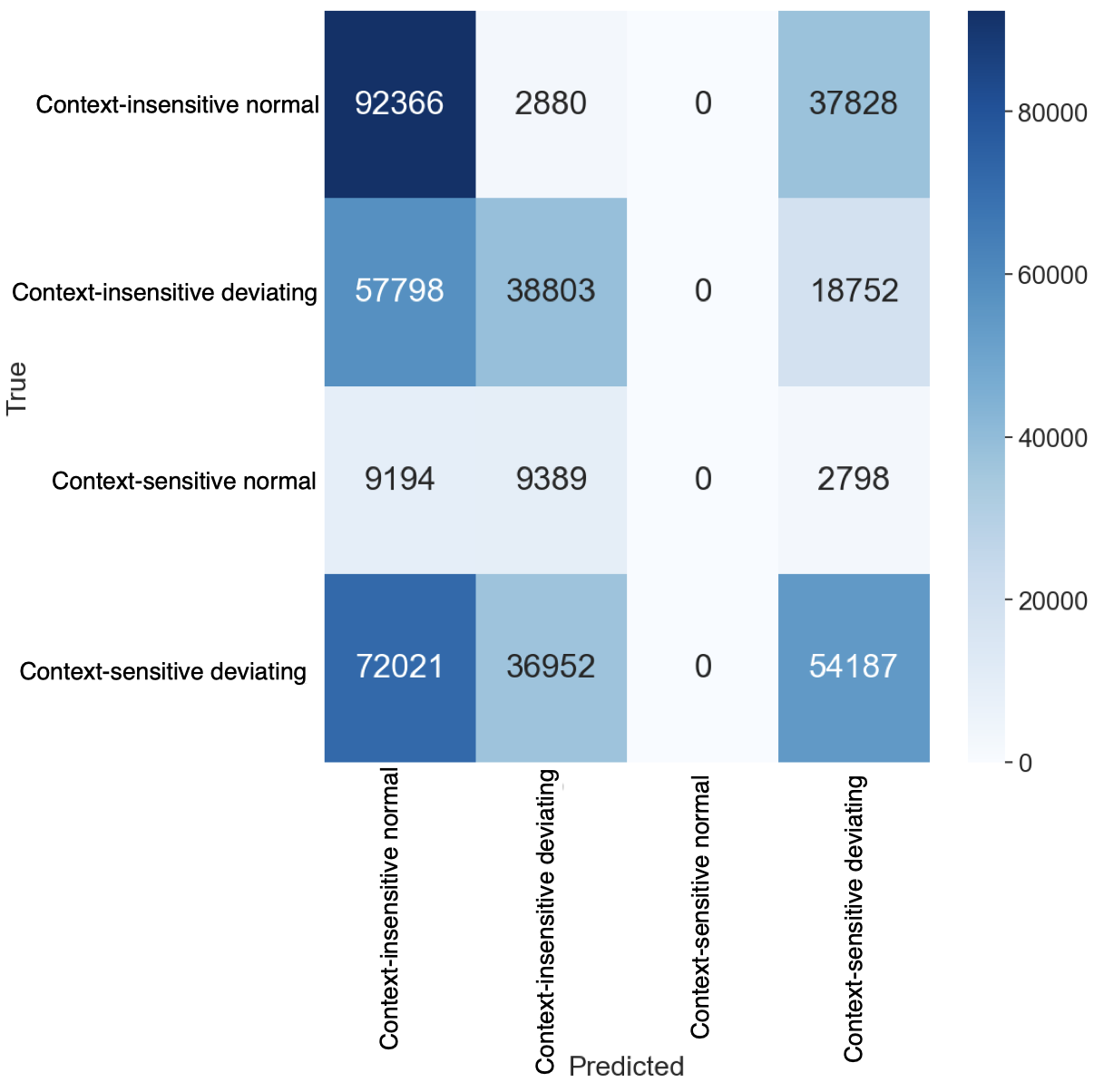}
        \caption{Inductive}
        \label{fig:cm-inductive}
    \end{subfigure}
    \begin{subfigure}{.5\textwidth}
        \center
        \includegraphics[width=\linewidth]{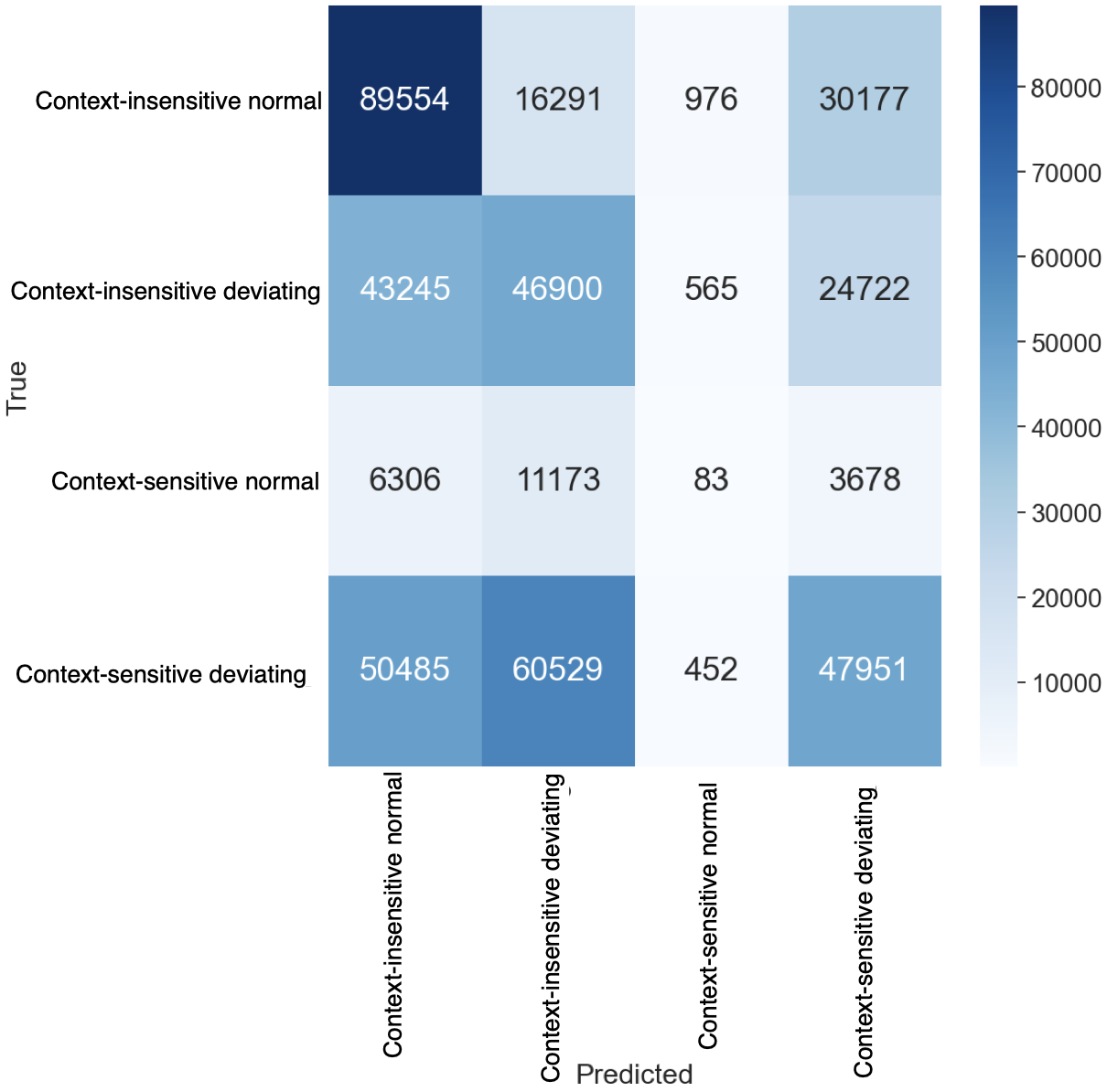}
        \caption{Autoencoder}
        \label{fig:cm-autoencoder}
    \end{subfigure}
    \caption{Confusion matrices summed over all 225 experiments of the respective context-aware deviation detection method}
    \label{fig:cm}
\end{figure}

Second, \autoref{fig:cm} shows two confusion matrices in \autoref{fig:cm} for \emph{Inductive} and \emph{Autoencoder}, summing the confusion matrix of each experiment.
The confusion matrix for \emph{Autoencoder} is representative for \emph{Profiles} and \emph{ADAR}, showing similar results.
The context-awareness generally improves the performance in all case studies by improving the detection of \emph{context-sensitive deviating} traces, but not by detection of \emph{context-sensitive normal} traces.
With respect to \emph{context-sensitive normal}, the framework's context-awareness has most of the time does not correctly predict the \emph{context-sensitive normal} traces (0 out of \emph{9,194} + \emph{9,389} + \emph{2,798} = \emph{21,381} \emph{context-sensitive normal} traces for \emph{Inductive} and \emph{83} out of \emph{6,306} + \emph{11,173} + \emph{83} + \emph{3,678} = \emph{21,240} traces for \emph{Autoencoder}). 
With respect to \emph{context-sensitive deviating}, the framework's context-awareness performs significantly better for the \emph{context-sensitive deviating} traces with \emph{54,186} of \emph{72,021} + \emph{36,952} + \emph{0} + \emph{54,187} = \emph{163,160} correctly predicted traces (\emph{Inductive}) and with \emph{47,951} of \emph{50,485} + \emph{60,529} + \emph{452} + \emph{47,951} = \emph{159,417} correctly predicted traces (\emph{Autoencoder}). 

\section{Conclusion}\label{sec:conclusion}
In this paper, we proposed a framework to support context-aware deviation detection.
The proposed framework can incorporate any existing unsupervised deviation detection methods with varying strengths and weaknesses and enhance them with various contextual aspects.
We have implemented the framework as an extensible web service with a dedicated user interface.
Moreover, we have evaluated the effectiveness of the framework by conducting experiments using representative deviation detection methods in different contextual scenarios.

This work has several limitations.
First, the proposed framework introduces several parameters that possibly affect the detection results, e.g., the negative and positive degree of $\mi{post}$ function, the threshold of $\mi{score}$ function, etc.
Second, the framework is dependent on the performance of the deviation detection method.
Third, using an event log as the input, the framework only indirectly measures external contexts.

Besides addressing the above limitations, in future work, we plan to extend the framework to support the root cause analysis of context-aware deviations.
We can analyze the relevant context of context-aware deviating instances and trace back the relevant context measure, e.g., high workload.
Moreover, we plan to extend the framework to consider contexts of different time window lengths, e.g., context in \textit{week}, \textit{day}, and \textit{hour}.
Another direction of future work is to develop different post functions to improve the performance of the context-aware deviations.

%
%
%
\bibliographystyle{splncs04}
\bibliography{references}

\begin{thebibliography}{10}
\providecommand{\url}[1]{\texttt{#1}}
\providecommand{\urlprefix}{URL }
\providecommand{\doi}[1]{https://doi.org/#1}

\bibitem{vanderaalst2012alignments}
van~der Aalst, W.M.P., Adriansyah, A., van Dongen, B.F.: Replaying history on
  process models for conformance checking and performance analysis. Wiley
  Interdiscip. Rev. Data Min. Knowl. Discov.  \textbf{2}(2),  182--192 (2012)

\bibitem{vanderaalst2012context}
van~der Aalst, W.M.P., {Dustdar}, S.: Process mining put into context. IEEE
  Internet Computing  \textbf{16}(1),  82--86 (2012)

\bibitem{Bezerra2013deviationsimplealgorithms}
Bezerra, F., Wainer, J.: {Algorithms for anomaly detection of traces in logs of
  process aware information systems}. Information Systems  \textbf{38}(1),
  33--44 (2013)

\bibitem{Bohmer2017deviationreview}
B{\"{o}}hmer, K., Rinderle{-}Ma, S.: Anomaly detection in business process
  runtime behavior - challenges and limitations. CoRR  \textbf{abs/1705.06659}
  (2017)

\bibitem{Bohmer2017deviationwindowbasedsignatureextraction}
B{\"{o}}hmer, K., Rinderle{-}Ma, S.: Multi instance anomaly detection in
  business process executions. In: et~al, J.C. (ed.) {BPM} 2017. LNCS, vol.
  10445, pp. 77--93 (2017)

\bibitem{Bohmer2020deviationadar}
B{\"{o}}hmer, K., Rinderle-Ma, S.: {Mining association rules for anomaly
  detection in dynamic process runtime behavior and explaining the root cause
  to users}. Information Systems  \textbf{90},  101438 (2020)

\bibitem{costa2007contextrules}
Dockhorn~Costa, P., Almeida, J.P.A., Ferreira~Pires, L., van Sinderen, M.:
  Situation specification and realization in rule-based context-aware
  applications. In: Indulska, J., Raymond, K. (eds.) Distributed Applications
  and Interoperable Systems. pp. 32--47 (2007)

\bibitem{Jalali2010geneticdeviationclassicprocessmodelusinggeneticdiscovery}
Jalali, H., Baraani, A.: {Genetic-based anomaly detection in logs of process
  aware systems}. World Academy of Science, Engineering and Technology
  \textbf{64}(4),  304--309 (2010)

\bibitem{Kronsbein2014contextinternal-org-resource-customer-external}
Kronsbein, D., Meiser, D., Leyer, M.: {Conceptualisation of contextual factors
  for business process performance}. Lecture Notes in Engineering and Computer
  Science  \textbf{2210} (2014)

\bibitem{van2013imf}
Leemans, S.J.J., Fahland, D., van~der Aalst, W.M.P.: Discovering
  block-structured process models from event logs containing infrequent
  behaviour. In: et~al, N.L. (ed.) Business Process Management Workshops -
  {BPM} 2013 International Workshops. LNBIP, vol.~171, pp. 66--78 (2013)

\bibitem{Li2017profilesdeviation}
Li, G., {van der Aalst}, W.M.P.: {A framework for detecting deviations in
  complex event logs}. Intelligent Data Analysis  \textbf{21}(4),  759--779
  (2017)

\bibitem{Mannhardt2016alignmentsbalanceddeviation}
Mannhardt, F., de~Leoni, M., Reijers, H.A., van~der Aalst, W.M.P.: {Balanced
  multi-perspective checking of process conformance}. Computing
  \textbf{98}(4),  407--437 (2016)

\bibitem{Nolle2018autoencoderdeviation}
Nolle, T., Luettgen, S., Seeliger, A., M{\"{u}}hlh{\"{a}}user, M.: Analyzing
  business process anomalies using autoencoders. Mach. Learn.
  \textbf{107}(11),  1875--1893 (2018)

\bibitem{Nolle2019deviationrnn}
Nolle, T., Luettgen, S., Seeliger, A., M{\"{u}}hlh{\"{a}}user, M.: {BINet}:
  Multi-perspective business process anomaly classification. CoRR
  \textbf{abs/1902.03155} (2019)

\bibitem{Pauwels2019bayesiannetworksdeviaton}
Pauwels, S.: {An anomaly detection technique for business processes based on
  extended dynamic Bayesian networks}. Proceedings of the ACM Symposium on
  Applied Computing  \textbf{Part F1477},  494--501 (2019)

\bibitem{Song2019contextreview}
Song, R., Vanthienen, J., Cui, W., Wang, Y., Huang, L.: Towards a comprehensive
  understanding of the context concepts in context-aware business processes.
  In: Betz, S. (ed.) {S-BPM} {ONE} 2019. pp. 5:1--5:10 (2019)

\bibitem{warrender1999tstide}
Warrender, C., Forrest, S., Pearlmutter, B.A.: Detecting intrusions using
  system calls: Alternative data models. In: 1999 {IEEE} Symposium on Security
  and Privacy. pp. 133--145 (1999)

\bibitem{ye2012situationreview}
Ye, J., Dobson, S., McKeever, S.: Situation identification techniques in
  pervasive computing: A review. Pervasive and Mobile Computing  \textbf{8}(1),
   36--66 (2012)

\end{thebibliography}
\end{document}